\newif\ifcomments\commentsfalse
\newif\ifcompress\compresstrue

\documentclass{article}

\usepackage[final]{neurips_distshift_2023}

\usepackage{natbib}

\usepackage[utf8]{inputenc} %
\usepackage[T1]{fontenc}    %
\usepackage{hyperref}       %
\usepackage{url}            %
\usepackage{booktabs}       %
\usepackage{amsfonts}       %
\usepackage{amssymb}       %
\usepackage{nicefrac}       %
\usepackage{microtype}      %
\usepackage{xcolor}         %
\usepackage[pdftex]{graphicx}
\usepackage{paralist}
\usepackage{tikz}

\ifcomments
\newcommand{\warning}[1]{{\color{red}#1}}
\newcommand{\todo}[1]{{\color{red}#1}}
\newcommand{\comment}[1]{{\color{red}#1}}
\newcommand{\mh}[1]{{\color{blue}[MH: #1]}}
\newcommand{\jb}[1]{{\color{cyan}[JB: #1]}}
\newcommand{\todoa}[1]{\todo{[AG: #1]}}
\else
\newcommand{\warning}[1]{}
\newcommand{\todo}[1]{}
\newcommand{\comment}[1]{}
\newcommand{\mh}[1]{}
\newcommand{\jb}[1]{}
\newcommand{\todoa}[1]{}
\fi

\title{Revisiting Dynamic Evaluation: \\ Online Adaptation for Large Language Models}

\author{%
  Amal Rannen-Triki$^*$
  \And J\"org Bornschein\thanks{equal contribution}
  \And Razvan Pascanu
  \And Marcus Hutter
  \And Andr\'as Gy\"orgy
  \And Alexandre Galashov
  \And Yee Whye Teh
  \And Michalis K. Titsias \\
  \And \textnormal{Google DeepMind, London}
}

\begin{document}

\maketitle

\begin{abstract}
We consider the problem of online finetuning the parameters of a language model at test time, also known as dynamic evaluation. 
While it is generally known that this approach improves the overall predictive performance, 
especially when considering distributional shift between training and evaluation data, 
we here emphasize the perspective that online adaptation turns 
parameters into temporally changing states and provides a form of context-length extension with \emph{memory in weights}, more in line with the concept of \emph{memory} in neuroscience. 
We pay particular attention to the speed of adaptation (in terms of sample efficiency), sensitivity to the overall distributional drift, 
and the computational overhead for performing gradient computations and parameter updates. Our empirical study provides insights on when online adaptation is particularly interesting. 
We highlight that with online adaptation the conceptual distinction between in-context learning and finetuning blurs%
: both are methods to condition 
the model on previously observed tokens.

\end{abstract}

\mh{Do not delete comments, but comment them out when addressed. Make all invisible by setting comments switch to false}

\section{Introduction}
Transformer-based language models can be conceptualized as systems with two distinct memory components:
One is given by the model's parameters, and learning (through gradient descent) can be seen as encoding information from the training set into this memory. 
The other one is the context, which roughly correspond to the persistent hidden states in recurrent neural networks.
For transformers, context is a non-parametric
form of memory: the tokens within the attention window.
LLMs rely heavily on context to condition the model towards desired behaviour. %
However, the prompt is a precious resource, and for transformers the cost of inference grows with the size of the attention window.
This becomes more problematic in multimodal systems, where images or short videos can easily exhaust the context tokens we can afford to use.
With dynamic evaluation \cite{krause2018dynamic, krause2019dynamic}, the idea of updating model parameters at test time, model 
parameters become part of the temporal, changing state of the model. Parameters can capture longer-term information that 
exceed the length of the context window and are also suited to adapt to \emph{distributional changes} that exceed the in-context adaptability of the model. Online learning therefore can be seen as one particular type of memory, particularly suited to changes like style or topic, which appear to the model as a distribution shift in the observations. 
We here investigate various trade-offs when online-adapting transformer-based LLMs with gradient descent on long text sequences.

\section{Methods for SGD online adaptation}

In this section, we describe the methods that we put in place to make our study possible. These methods have two main goals: 
\begin{inparaenum}[(i)]
\item learn from a sequence of tokens longer than the model context, and
\item efficiently update the model parameters and reduce the memory and/or compute footprint of online adaptation. 
\end{inparaenum}
The challenge is that transformer implementations operate on a limited, typically fixed number of tokens each time they are invoked. To operate on longer sequences they have to be broken into sub-sequences, and the model implementation operates on such a sub-sequence at a time. We experiment with two strategies:

\textbf{Overlapping.} 
\todoa{Suggestion for the next sentence: Considering a transformer implementation that processes $2n$ tokens at a time with an overlap window of $n$ tokens, with the overlapping evaluation strategy we invoke the model on tokens $[0,2n)$ predicting tokens $[n,2n)$, then on $[n,3n)$ predicting $[2n,3n)$, then on $[2n,4n)$ predicting $[3n,4n)$, etc.}
Considering a transformer implementation that processes 1000 tokens at a time: Choosing an overlap of 500, we invoke the model on tokens $[0, 1000)$, $[500, 1500)$, $[1000, 2000)$, etc. 
Each token is encoded twice, once as a new token and once as context for later tokens.
For the purpose of computing the test-set log-losses, only the first encounter is recorded. 
For gradient steps however we let all tokens supplied during model invocation contribute to the loss and thus to the gradient computation. 
By adjusting the overlap we can adjust the computational cost but also the number of gradient steps performed on each token.

\textbf{Transformer-XL style.} 
In \citep{dai2019transformer} and \citep{krause2019dynamic} the authors use a form of \emph{streaming} or \emph{KV caching}: instead of attending only to tokens that are computed during the current forwards pass, the model can also attend to previously computed keys and values. As a result, each token is processed exactly once, and serves as a prediction target in exactly one gradient step. 

Empirically we observe that Transformer-XL-style adaptation performs as well as overlapping adaptation and requires significantly fewer computational resources; see Appendix~\ref{sec:overlapping}. We therefore focus on Transformer-XL-style adaptation. 

Compared to static evaluation, online adaptation requires additional computational resources for the backward pass and, typically, additional memory for the optimizer state. We  investigate two approaches to mitigate these costs: 

\textbf{Reducing the Update Frequency.} In order to vary the computational cost of online learning, and to construct Pareto fronts that highlight the compute vs.\ performance trade-offs, we update the parameters only every $n$th forward step. While this approach leads to a suboptimal performance, we show in our experiments that it can strike interesting trade-offs. 

\textbf{Online LoRA adaptation.} Reducing the memory footprint of finetuning LLMs has recently been an active research direction. A successful approach proposed in \cite{lora} consists of adding low-rank matrices to the transformer layers, %
and only adapting the parameters of these matrices during finetuning. This greatly reduces the number of parameters that need to be adapted, and therefore the memory requirements for downstream adaptation with negligible additional computational cost.

\section{Experimental Setup}

To investigate the online adaptability of transformers we use the books from Project-Gutenberg 
(the PG-19 dataset \citep{raecompressive2019}) as a source for long and consistent text sequences.
The starting point for all experiments is a transformer pretrained on the C4 dataset \citep{C4}.
We experiment with different model-sizes, between 150M and 1B parameters; 
Appendix \ref{sec:details} contains the experimental details. The pretrained models are then finetuned on books from the PG-19 
training set because the content and style of text in the C4 dataset, which consists mostly of internet-scraped data, and PG-19, which contains $28'602$ %
books\footnote{PG-19 training set size} from before 1919, is significantly different
(see Figure \ref{fig:scaling} and Appendix~\ref{sec:dataset}).
The finetuned models are then tested against the sequence of 100 books from the PG-19 test set.
For the test sequence we record the (cumulative) log-losses for all tokens.
We concatenate the 100 books from the PG-19 test set, in the order they are stored, to form a fixed sequence of 11.8M tokens.

\begin{figure}[t]
  \includegraphics[width=0.45\linewidth]{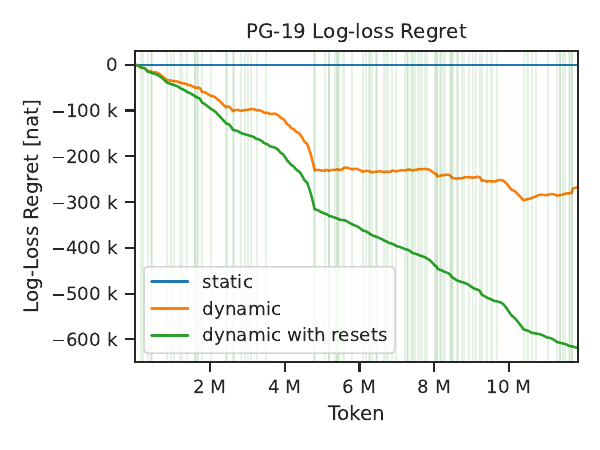}
  \includegraphics[width=0.45\linewidth]{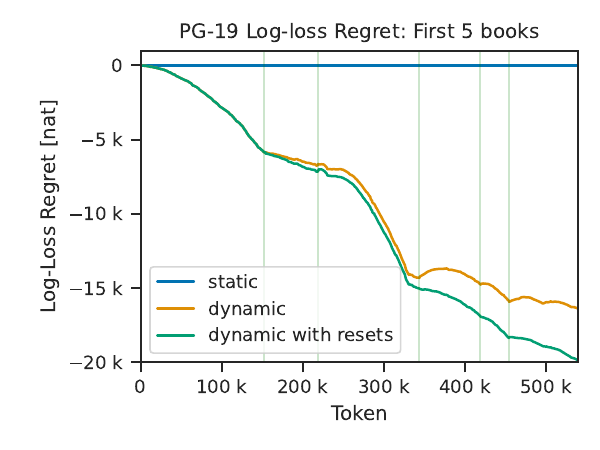} \\
  \vspace{-0.6cm}
  \caption{
  {\bf Left:} Cumulative log-loss for dynamic evaluation relative to static evaluation (regret). The starting point is always a model that has been finetuned on the PG-19 training set. 
  {\bf Right:} Detailed view of the regret for the first 5 books.
  Vertical green lines indicate the beginning of new books.
  }
  \label{fig:regret}
\end{figure}

Figure \ref{fig:regret} visualizes typical results: We compare static evaluation vs.\ dynamic evaluation vs.\ dynamic evaluation where the model is reset to the finetuned model at each book boundary.
The static model accumulates in total 26.73 M nats log-loss on the test sequence (2.26 nats/token), while the dynamic models accumulate 26.38 M and 26.20 M nats respectively (corresponding to 2.23 and 2.20 nat/token). 
The regret plots show the cumulative log-loss relative to the static comparator: a flat curve indicates that a model has on average the same per-token log-loss as the comparator around the position, 
while positive and negative slopes indicate a locally higher or lower log-loss repspectively.
It often takes the online adapting model some thousand tokens to show a clear advantage over the static model, and just after book boundaries, a continuously adapted model often underperforms compared to the pretrained (static or dynamic) model.
This may not be surprising as the continuously learning model has specialized to the previous books, but also suggests that more advanced adaptation methods should be able to close this gap.
\mh{Without finetinung on the PG-19 training set, that should be the converse, right? Because it's still learning the PG-19 distribution,
and resetting doesn't. If true that would mean there is a reversal at some point: initially resetting harms, but eventually resetting helps.
Worth to mention? So maybe ``This may not be surprising'' should be replaced by ``A prior it is unclear which should be better''?
}

\subsection{Overlapping vs. Transformer-XL style updates}
\label{sec:overlapping}

\begin{figure}[h!]
    \centering
    \includegraphics[width=0.6\linewidth]{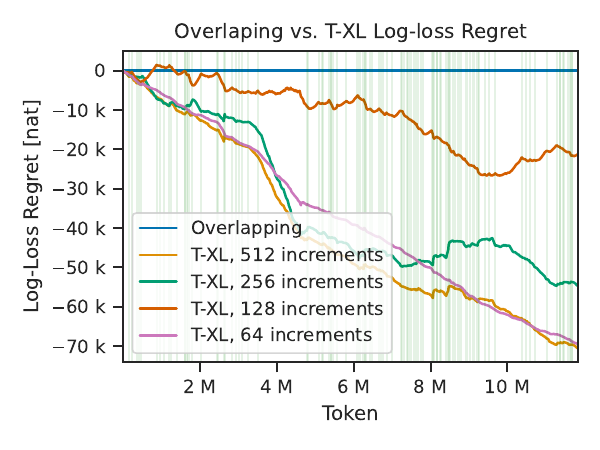}
    \caption{
    Regret plot of Transformer-XL style online learning with varying increment-size relative to 
    Overlapping online learning with 0.5 overlap. We observe that Transformer-XL style online learning generally leads to 20k to 70k fewer accumulated loss. However, 70k nats
    over 11.8M tokens corresponds to only about 0.006 nat/token uplift -- a minuscule
    improvement compared to the differences plotted in Figures \ref{fig:regret} to \ref{fig:lora_regret}.
    }
    \label{fig:overlapping}
\end{figure}

Figure \ref{fig:overlapping} shows the results when using Transformer-XL style online 
adaptation with varying increment lengths relative to overlapping online learning with 50\% (= 1024 tokens)
overlap. We observe that the differences are minuscule. We thus focus on Transformer-XL style online
learning because it has the computational advantage of encoding every token only once.

\subsection{Online Learning - An analysis of compute vs.\ performance}

In order to understand the impact of online learning and its interaction with in-context learning, we conduct a large exploration varying: 
\begin{itemize}
    \item \textbf{The number of samples used for finetuning:} This allows us to control the distribution shift that the model is faced with --  the more the model is finetuned on the training set of PG-19, the more it is \emph{in-distribution} with respect to the test sequence.
    \item \textbf{The model context size:} We hypothesize that online learning can have similar benefits as larger context windows. We therefore compare online adapted models with short context windows to in-context learning in models with longer context windows. 
    \item \textbf{The model size:} We are interested in understanding how our observations generalize across different scales. 
\end{itemize}

\begin{figure}
  \includegraphics[width=0.95\linewidth]{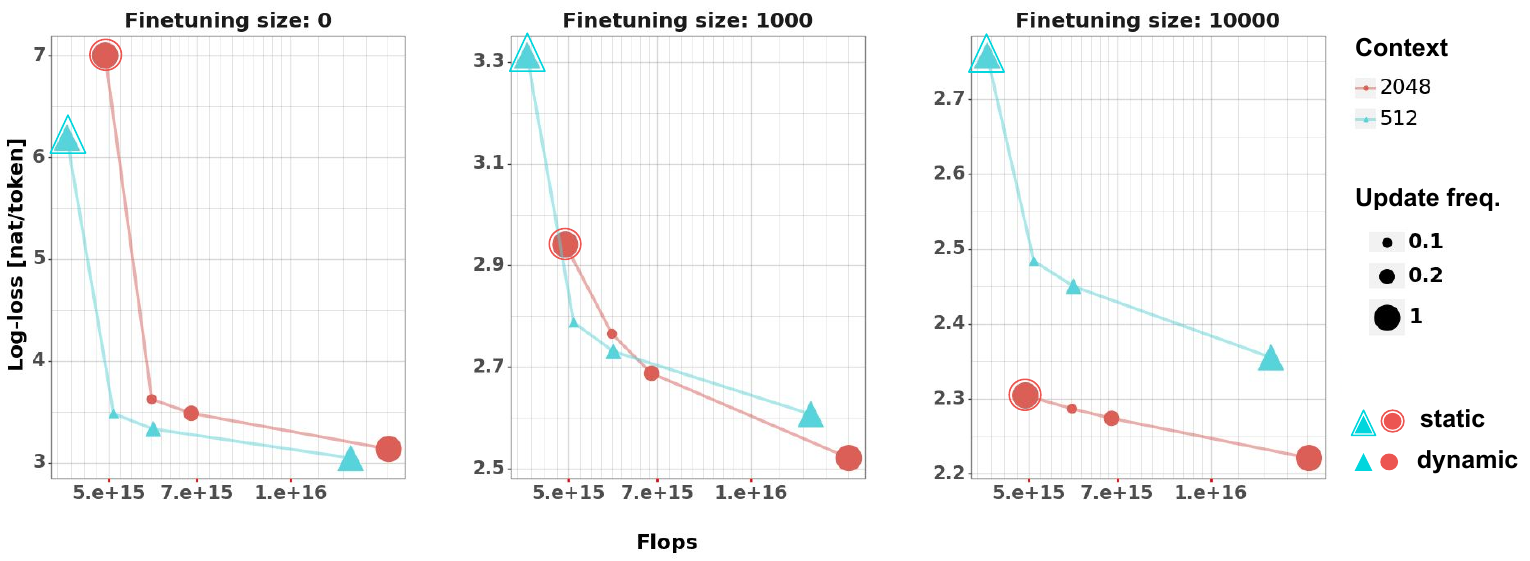}
  \ifcompress\vspace{-0.3cm}\fi %
  \caption{
  {\bf Performance vs.\ compute (FLOPs) for static \emph{and} dynamic evaluation.} Models with 1 billion parameters, varying the context size and the number of finetuning samples (books). The Pareto front is constructed by varying the update frequency. 
  \label{fig:1b-context}
  \todoa{The update frequency as never defined exactly. Amal will define}
  }
\end{figure}

In Figure~\ref{fig:1b-context}, we first look at a single model size (1B parameters) and compare the Pareto fronts corresponding to two context sizes (512 and 2048) as we vary the number of samples the model is finetuned on. We observe that when the model is directly updated online on the PG-19 test set without prior finetuning, the models with a smaller context exhibit a better compute to performance trade-off than the models with a larger context (left panel). As the amount of finetuning increases, this advantage is reduced (middle panel) and even inverted (right panel). These observations generalize to other model sizes (figures in the appendix). These results suggest that the models favor \emph{memory in weights} when faced with a large distribution shift between the pretraining and online adaptation data. When the models are more in-distribution with respect to the online data, the results suggest that at a fixed budget, it is better to use a model with a larger context window. Online adaptation however unsurprisingly always improves the performance of the models. Moreover, models with shorter context and online adaptation can achieve a competitive performance to the models with longer context. While this can be more expensive in terms of FLOPs, it comes at a lower memory requirement.

\begin{figure}
  \includegraphics[width=0.95\linewidth]{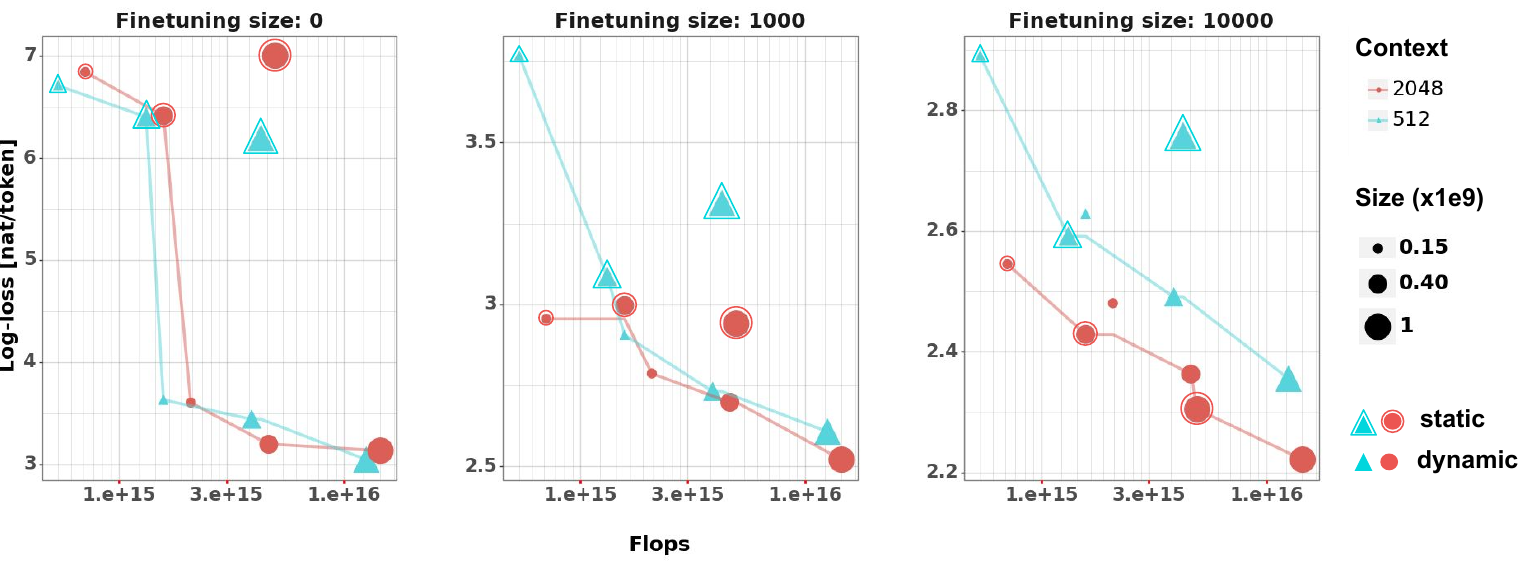}
  \ifcompress\vspace{-0.3cm}\fi %
  \caption{
  {\bf Performance vs.\ compute (FLOPs) for static \emph{and} dynamic evaluation.} Varying model and context sizes, and the number of finetuning samples. The models are updated with every new observation. \label{fig:compare-model-sizes}
  \mh{Depending on the restrictions flops memory ... different trade-offs}
  }
\end{figure}

Figure~\ref{fig:compare-model-sizes} gathers the results obtained with different model sizes (150M , 400M and 1B parameters).\mh{inconsistent with Figure 3 [Amal] Figure 3 does gather results from different model sizes. MH: Yes, but Fig.3 says sizes are 200M,400M,600M,800M,1B.}
For this figure, we show only models that are updated at every step (corresponding to an update frequency of 1 in Figure~\ref{fig:1b-context}). We observe that when we increase the amount of finetuning, the number of the static models that appear on the Pareto front increases. We however similarly observe that online adaptation always improves the performance, smaller models with online adaptation can achieve a competitive and sometimes better performance than larger models. 
\todoa{Isn't the take-home message the following: Online learning comes with a fixed computational cost (depending on the model size, including the context length) and performance improvements. The performance improvement is more when the initial model is worse (in other words, more out of distribution) and becomes less significant when the initial model is better. As a result, more static models appear on the Pareto front when the latter happens, which is exactly when we do more finetuning. MH: Amal will take Andras version.}

\begin{figure}[h!]
  \includegraphics[width=0.5\linewidth]{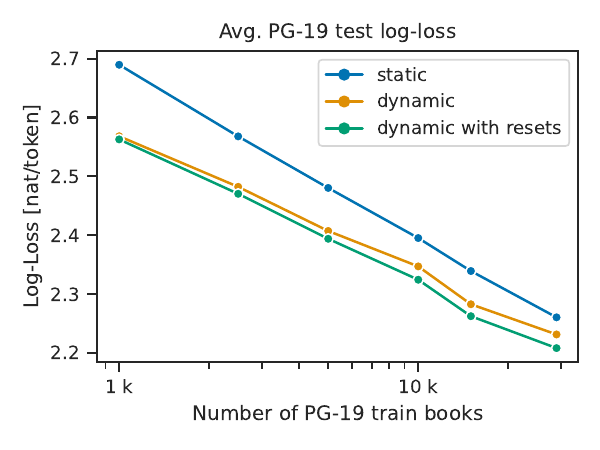}
  \includegraphics[width=0.5\linewidth]{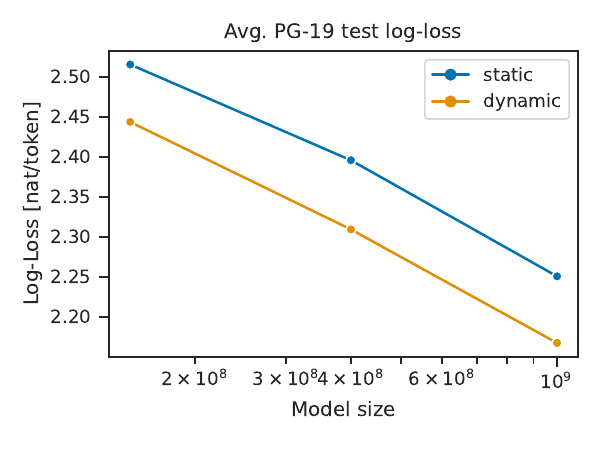}
  \vspace{-0.7cm}
  \caption{
  {\bf Left:} Scaling of the average PG-19 test loss with size of PG-19 i.i.d. finetuning dataset (for the 400 M model).
  {\bf Right:} Average test NLL as a function of the model size (after finetuning on 10k books) 
  \label{fig:scaling}}
\end{figure}

Finally, Figure~\ref{fig:scaling} shows how the static and dynamic model performance scale with the distirbution shift (amount of finetuning data) and the model size. 
The figure shows that while finetuning reduces the gap between static and online evaluation, this gap does not disappear, but rather becomes constant once the model adapts to the change in distribution.
Similarly, increasing model size, while also improving overall performance, does not replace the benefit of online learning.

\section{Conclusion}

In this paper, we offered a new perspective of online adaptation of LLMs (a.k.a.\ dynamic evaluation).
Through extensive experiments, we show that when a model is faced with a significant distribution shift, online learning with a smaller context window and/or a smaller model can lead to a better compute-performance Pareto front, which suggests a superiority of the \emph{memory in weights} over the \emph{memory in activations} associated with in-context learning. This advantage is reduced and eventually disappears when the model is finetuned to the target distribution before the online adaptation phase. We however observe (as is classically the case with dynamic evaluation) that online learning always improves the performance. Moreover, for models that are in-distribution, online learning with a smaller context (and therefore smaller memory requirements) can close the gap with in-context learning with a larger context. It is also worth noting that the best results observed with online adaptation employ a simple strategy to avoid that the models overfit to data in the local context by resetting the weights to their value at the start of the online adaptation phase. 

This study opens up many interesting research avenues, such as improving the efficiency of online learning (in terms of memory or compute),
automatic detection of reset points, and a better understanding of the difference between what the weight memory and the activation memory capture. In particular, it is to be expected that weight memory would be better suited to store the style or topic of a discussion, which are perceived by the model as a \emph{distribution change}. Context and retrieval is better at capturing details, leading to a separation of concerns similar to how memory of biological systems is categorized into different types in cognitive sciences or neuroscience.

\medskip
\bibliographystyle{abbrvnat}

\bibliography{main}

\newpage
\section*{Supplementary Material}
\renewcommand{\thesubsection}{\Alph{subsection}}

\subsection{Adaptation dataset choice}
\label{sec:dataset}
\begin{figure}[h!]
    \centering
    \includegraphics[width=0.9\textwidth]{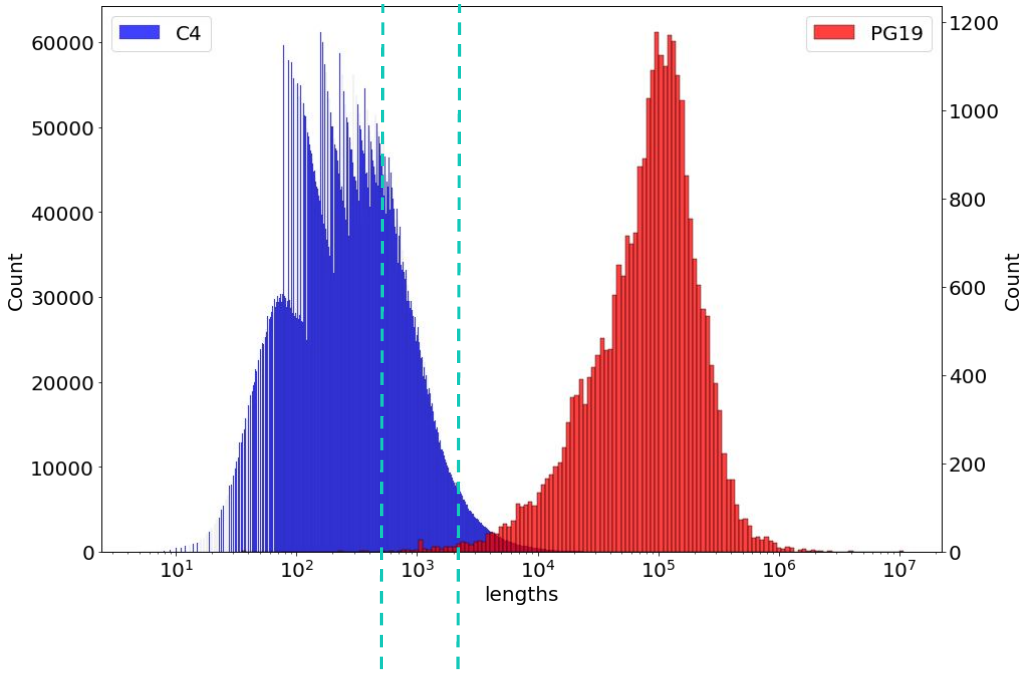}
    \caption{Comparing sequence length distribution between PG-19 and C4 datasets. For C4, this histogram corresponds to a subset of 10 million samples. The vertical dashed lines show the context length used in our experiments.\mh{Why are there only very few (less than 100) red lines. If these \emph{are} 28'602 lines,
    then convert this to a proper histogram. For $n$ data points, a histogram between $n^{1/3}$ and $n^{1/2}$ bins is ideal, so 100 bins is probably good, of course on an exponential spacing since the x-axis is logarithmic.}\mh{Font size is way too small} 
    }
    \label{fig:seq-length}
\end{figure}

\begin{figure}[h!]
    \centering
    \includegraphics[width=0.8\textwidth]{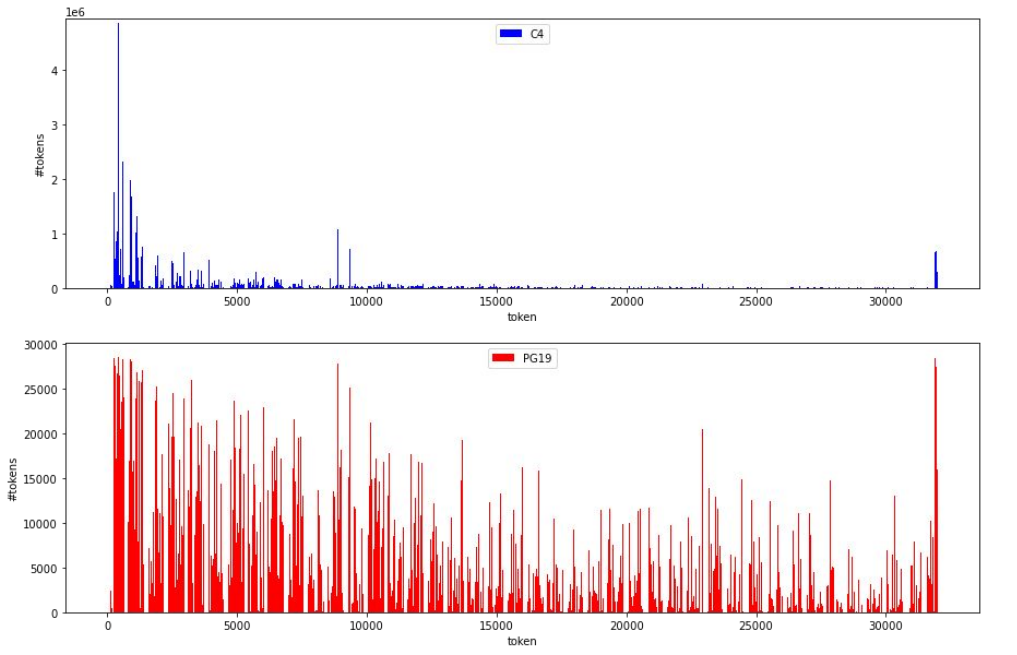}
    \caption{Comparing token distribution between PG-19 and C4 datasets. For C4, this histogram represents a subset of 10 million samples.\mh{How are the tokens sorted? Makes little sense to use the arbitrary numerical order. If you sort them wrt decreasing frequency in C4 and use this same sorting for C4 and PG19, you get much more meaningful plots. Once done, you probably want/need to plot y=log(\#tokens) and possibly even x=log(token)}}
    \label{fig:token-dist}
\end{figure}

In order to test the role of online adaptation as a context extension mechanism, the criteria we used to select a downstream dataset are the following: 
\begin{itemize}
    \item \textbf{Sequence lengths:} Our hypothesis is that online learning extends the context beyond the window with which the transformer is trained. In order to test this, we need a dataset that has a large number of entries that exceed our models' context length. Figure~\ref{fig:seq-length} shows the length distribution in PG-19 and in 10 million samples of C4. PG-19 contains $28'602$ books. We observe that over $99\%$ of the PG-19 dataset exceeds the largest context that we use in our experiments, compared to only $3\%$ for C4.
    
    \item \textbf{Token distribution:} Our study considers online learning when the model is faced with different levels of distribution shift. We therefore chose a pretraining and an online dataset that have different distributions. Figure~\ref{fig:token-dist} shows the token distribution for C4 and PG-19, confirming that they are good candidates for our study. In our experiments, to vary the distribution shift, the pretrained model on C4 is first finetuned on a subset of PG-19 (of size $0$, $1'000$ and $10'000$) before going through the phase of online adaptation. 
\end{itemize}

\subsection{Experimental Details}
\label{sec:details}

We use standard transformer architectures \citep{vaswani2017attention} with pre-layer normalization \citep{xiong2020layer}, GeLU activation functions and relative positional encoding. The context size (attention-window) is 2048 tokens by default, however we additionally trained models with 512 %
attention size for ablation studies. 
We use a sentence piece tokenizer~\citep{kudo2018sentencepiece} with a vocabulary of 32k entries. The table details the architecture choices the for the three model sizes under consideration: 

\begin{center}
  \begin{tabular}{ccccc}
    \toprule
    model size & num blocks & backbone width & num heads & key/value size \\
    \midrule
    150M & 12 & 896 & 16 & 64 \\
    400M & 12 & 1536 & 12 & 128 \\
    1B & 24 & 2048 & 16 & 128 \\
    \bottomrule
  \end{tabular}
\end{center}

\paragraph{Pretraining.} Training on C4 is performed with AdamW \citep{loshchilov2017decoupled} with linear learning rate warm-up over 10k steps and successive cosine learning rate decay:

\begin{center}
  \begin{tabular}{ccccc}
    \toprule
    model size & training steps & batch size & max LR & weight decay \\
    \midrule
    150M & 12k & 128 & 2e-4 & 0.1  \\
    400M & 30k & 128 & 2e-4 & 0.1  \\
    1B & 48k & 256 & 2e-4 & 0.1  \\
    \bottomrule
  \end{tabular}
\end{center}

\paragraph{Finetuning.} We finetune the C4 pretrained model on books from the PG-19 training set to align it with the general distribution of text found in PG-19. 
By varying the size of the finetuning dataset we can adjust the model to be more or less in-distribution with respect to the evaluation data. We use AdamW \citep{loshchilov2017decoupled} for finetuning and maintain a separate validation set for early stopping and hyperparameter tuning. We use linear-warmup and cosine decay learning rate 
schedule as during pretraining, however we stop early as soon as the validation performance 
degrades. We sweep over [1e-4, 2e-4, 3e-4] as maximum learning rates and configure cosine learning rate decay to 
complete after 2 epochs. Stopping early and avoiding overfitting during the finetuning phase is crucial 
to obtain good performance during evaluation.
Note that finetuning, just like pretraining, is performed by i.i.d. sampling text segments from their respective training corpus.

\paragraph{Evaluation.} During the dynamic evaluation phase we again use AdamW and sweep over the learning rates [1e-6, 3e-6, 1e-5, 3e-5] without any learning-rate schedule. 
We always report and plot the results for the best performing learning-rate.
For Transformer-XL style updating we experimented with different 
{\em increment-sizes} $\in \{32, 64, 128, 256, 512\}$
(the number of new tokens processed in each forward pass) 
and observed that the results are not very sensitive to this choice (see Figure \ref{fig:overlapping}). 
If not mentioned otherwise, we  choose 128 for computational convenience. 
Note that each token can attend to the full 2048 previously computed keys and values.

\subsection{Online Learning - An analysis of compute vs.\ performance - More results.}
\begin{figure}[h!]
    \centering
    \includegraphics[width=\textwidth]{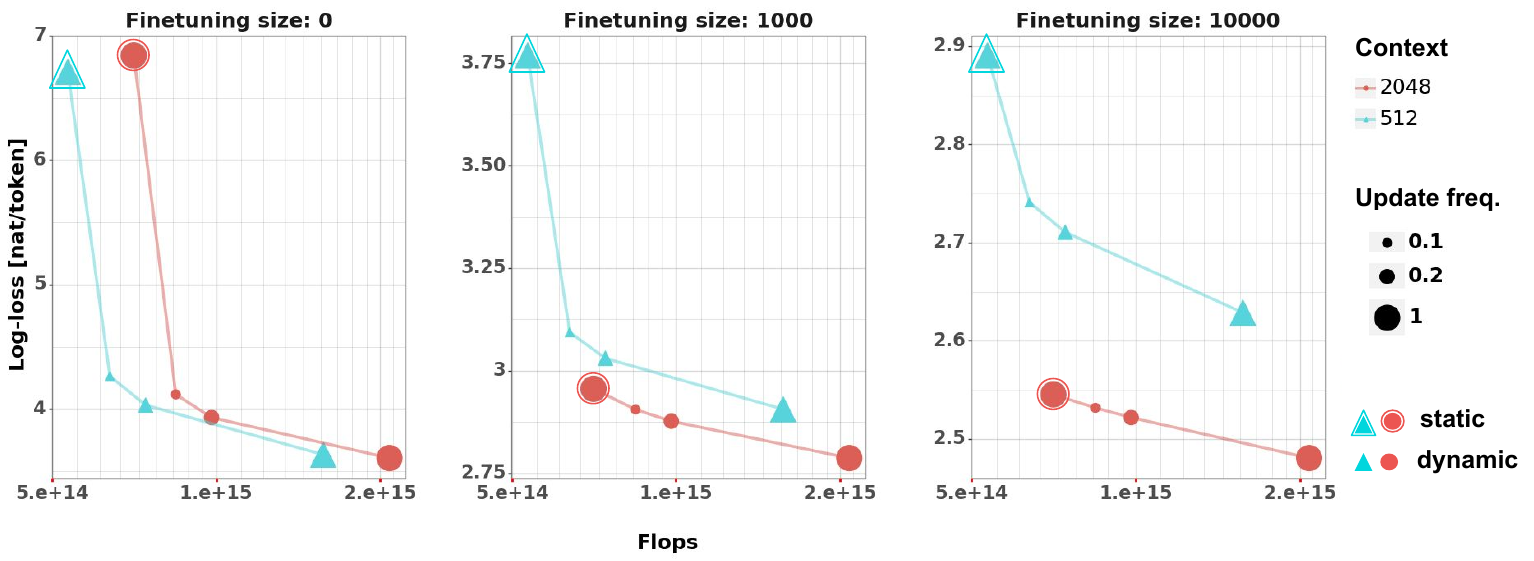}
    \caption{{\bf Performance vs.\ compute (FLOPs).} Models with 150 million parameters. Pareto fronts for each context size and number of finetuning samples constructed by varying the update frequency and evaluation methods (static vs.\ dynamic)}
    \label{fig:150m-context}
\end{figure}

\begin{figure}[h!]
    \centering
    \includegraphics[width=\textwidth]{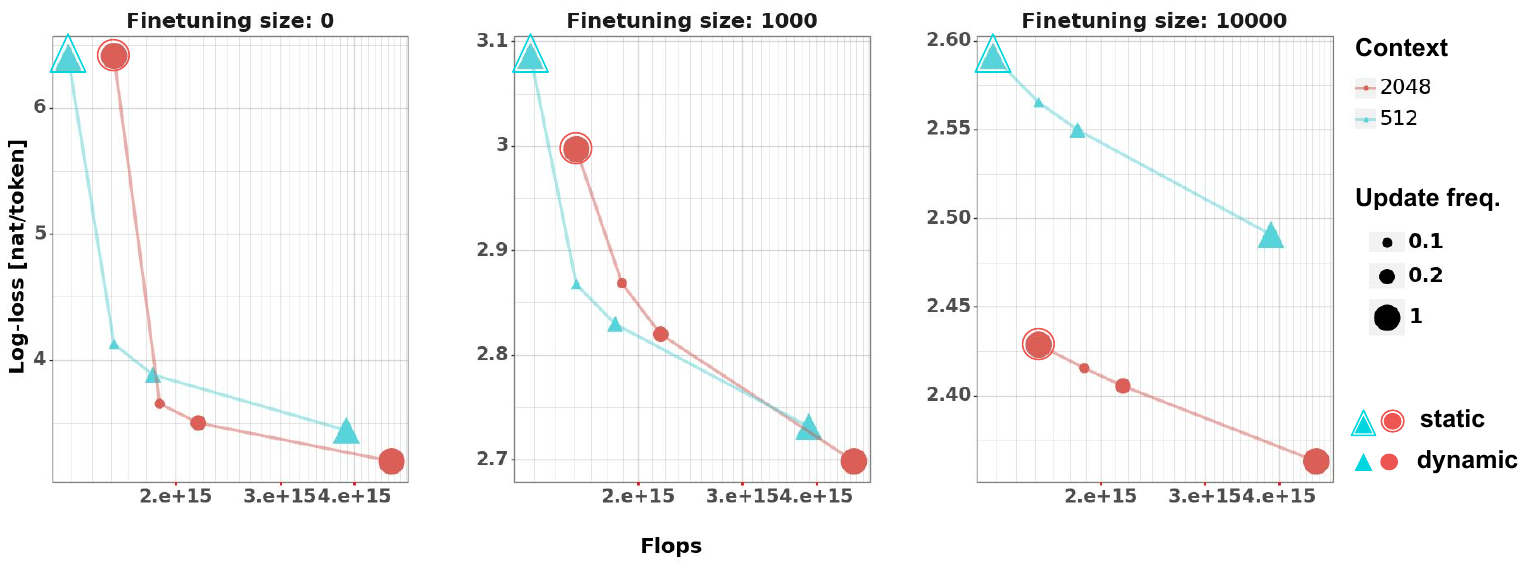}
    \caption{{\bf Performance vs.\ compute (FLOPs).} Models with 400 million parameters. Pareto fronts for each context size and  number of finetuning samples constructed by varying the update frequency and evaluation methods (static vs.\ dynamic)}
    \label{fig:400m-context}
\end{figure}

In Figure~\ref{fig:1b-context}, we report the performance-compute trade-off we observed with a 1B parameter model when we vary the context length.\mh{Just an observation: Fig.\ref{fig:compare-model-sizes} reports 200M,400M,...1B} In this section, we show results obtained with models of 150M (Figure~\ref{fig:150m-context}) and 400M (Figure~\ref{fig:400m-context}) parameters. We observe similar behavior across all model sizes: The models with  shorter context windows exhibit a competitive or better Pareto front than the models with longer context when they are adapted online on PG-19 without finetuning. The order of Pareto fronts is reversed when the models are first finetuned on the training data of PG-19. 

\subsection{Updating only subsets of the parameters}
\label{sec:subset}

\begin{figure}[h!]
  \centering
  \includegraphics[width=0.7\linewidth]{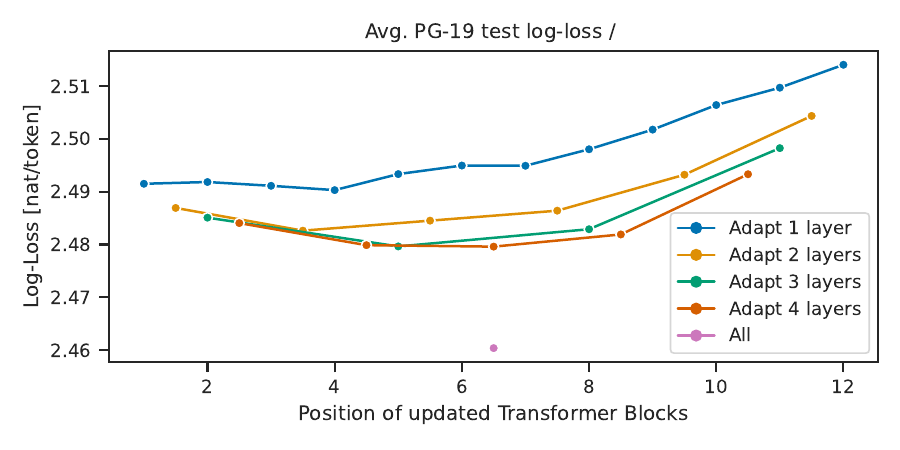}
  \vspace{-0.5cm}
  \caption{
  {\bf Adapting only a subset of the transformer blocks}:
  The blue curve shows the average log-loss when updating a single transformer block 
  at one of the layer indices 1 to 12. %
  The yellow curve shows the performance when updating two successive blocks at 
  layers 1+2, 3+4, 5+6, \dots.
  \label{fig:subset}}
\end{figure}

We explore the performance of dynamic evaluation when adapting only subsets of the complete model. 
For example, we can train only the top-most transformer block, a block in the middle, or at the beginning. 
Figure \ref{fig:subset} summarizes the results: We observe that generally adapting transformer blocks in the middle
of the stack is most effective.

\subsection{LoRA adaptation}\label{sec:lora}

In this section, we show the results that we obtained when applying LoRA~\cite{lora} to the MLPs in the 1B-parameter transformer. This is different from the way this adaptation strategy has been implemented in~\cite{lora} where LoRA is applied to the attention layers. The reason for this choice is emprirical: We applied adapters to multiple types of layers in the models, and adapting on the MLPs gave the best performance. 

\begin{figure}[h!]
    \centering
    \includegraphics[width=0.32\textwidth]{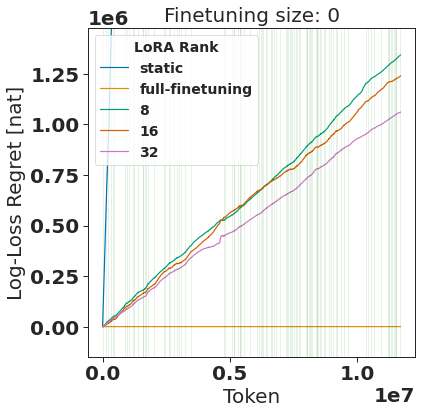}
    \includegraphics[width=0.32\textwidth]{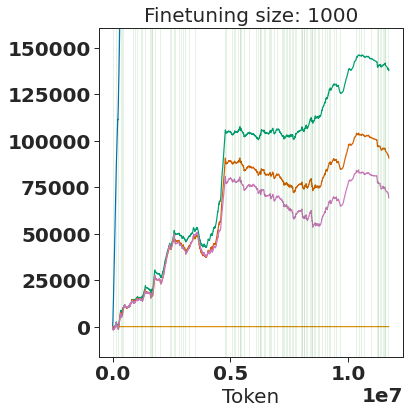}
    \includegraphics[width=0.32\textwidth]{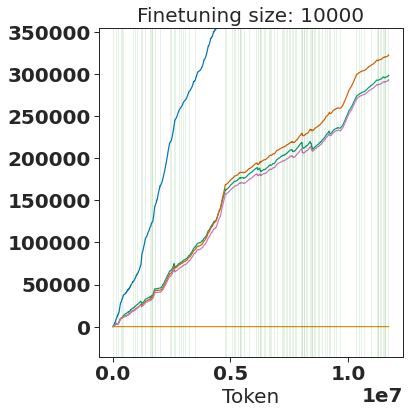}
    \caption{{\bf Cumulative log-loss for dynamic evaluation with LoRA relative to full finetuning (regret)}.
    Note the steep blue curves for static model evaluation: static evaluation quickly accumulates dramatically more log-loss 
    than all the dynamic evaluated models.\mh{Font size is still too small. Use vector pdf graphics, not pixel graphics. make legend 100\% transparent and no frame or have one legend outside the figure (all 3 legends are the same). Why is rank in random order?} }
    \label{fig:lora_regret}
\end{figure}

\begin{figure}[t!]
    \centering
    \includegraphics[width=\textwidth]{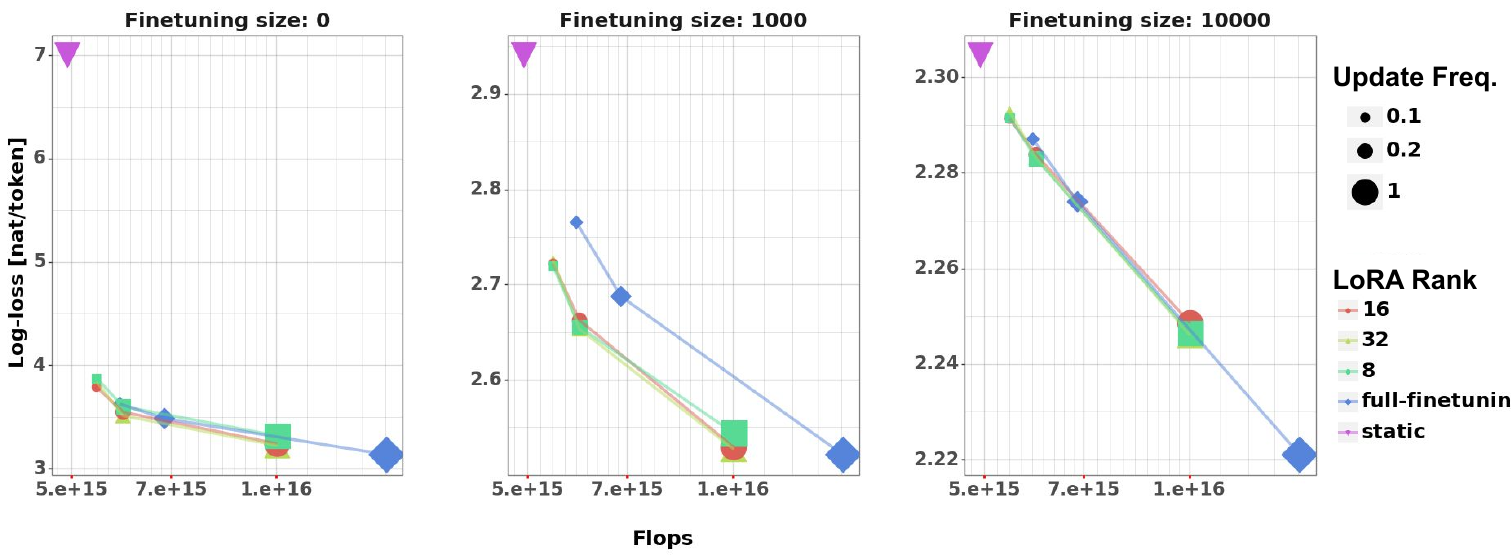}
    \caption{{\bf Performance vs.\ Compute (FLOPs)} Models with 1 billion parameters adapted online with LoRA (rank indicated in the legend) compared to static models and fully finetuned models). 
    }
    \label{fig:lora_flops}
\end{figure}

\begin{figure}[h!]
    \centering
    \includegraphics[width=\textwidth]{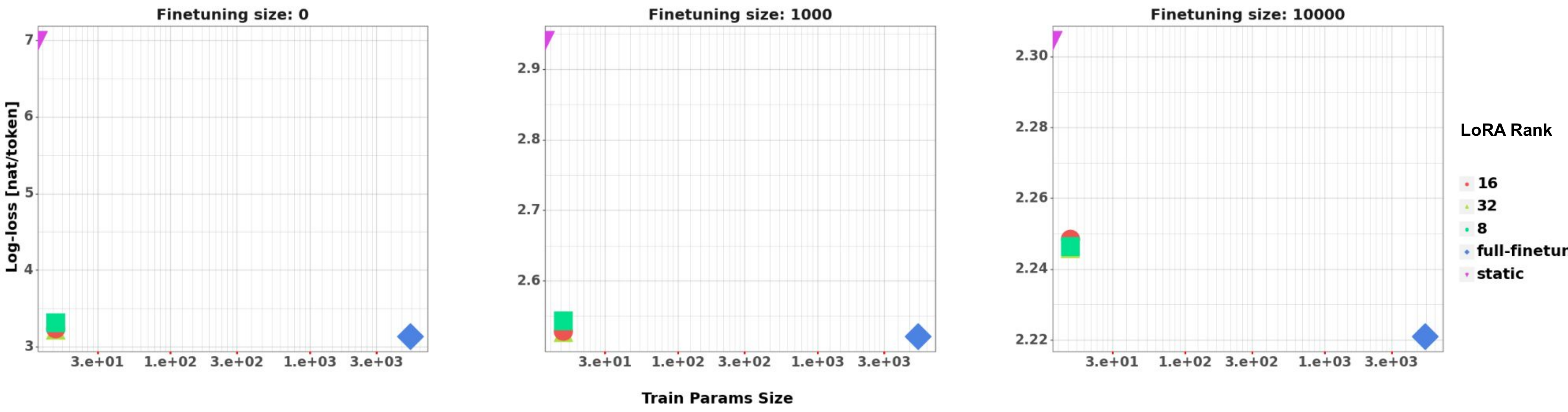}
    \caption{{\bf Performance vs.\ size of trainable parameters.} Models with 1 billion parameters adapted online with LoRA (rank indicated in the legend) compared to static models and fully finetuned models (LoRA rank = 0).}
    \label{fig:lora_size}
\end{figure}

Figure~\ref{fig:lora_regret} shows the cumulative log-loss for dynamic evaluation with LoRA relative to full finetuning. This regret plot shows that LoRA models achieve a lower performance than the fully finetuned model, but a significantly higher performance than the static model (blue line in the figure). It is worth noticing however that LoRA comes at a lower computational cost (as shown in Figure~\ref{fig:lora_flops}) and a much lower memory requirements for trainable parameters (as shown in Figure~\ref{fig:lora_size}).  

\end{document}